\documentclass[a4paper, 11pt, english]{article}

\usepackage{amsfonts}
\usepackage{amssymb}  
\usepackage{amsthm}
\usepackage{amsmath}
\usepackage{algorithmic}
\usepackage{courier} 
\usepackage{algorithm}
\usepackage{helvet}  
\usepackage{times}  
\usepackage{url}  
\usepackage{graphicx}
\usepackage{caption}
\usepackage{subcaption}

\graphicspath{{figures/}}
\newcommand{\ee}{{\rm e}\hspace{1pt}}

\newcommand{\norm}[1]{\Vert #1 \Vert}
\newcommand{\abs}[1]{\left| #1 \right|}
\def\phi{\varphi}

\renewcommand{\phi}{\varphi}

\newtheorem{thm}{Theorem}

\newtheorem{dfn}{Definition}

\begin{document}

\title{Learning Rate Adaptation for Federated and Differentially Private Learning}

\author{Antti Koskela$^1$ and Antti Honkela$^{2,3}$\\
  $^1$ Helsinki Institute for Information Technology HIIT, \\ 
  Department of Mathematics and Statistics, University of Helsinki, Finland \\
  $^2$ Helsinki Institute for Information Technology HIIT, \\
  Department of  of Computer Science , University of Helsinki, Finland\\
  $^3$ Department of Public Health, University of Helsinki, Finland}

\date{}

\maketitle
\begin{abstract}
  We propose an algorithm for the adaptation of the learning rate for 
  stochastic gradient descent (SGD) that avoids the need for
  validation set use. The idea for the adaptiveness comes from the technique of extrapolation: 
  to get an estimate for the error against the gradient flow which underlies SGD,  
  we compare the result obtained by one full step and two half-steps. 
  The algorithm is applied in two separate frameworks: federated and differentially private learning.
  Using examples of deep neural networks we empirically show that the adaptive algorithm
  is competitive with manually tuned commonly used
  optimisation methods for differentially privately training. We also show that it
  works robustly in the case of federated learning unlike commonly used optimisation methods.
\end{abstract}

%%%%%%%%%%%%%%%%%%%%%%%%%%%%%%%%%%%%%%%%%%%%%%%%%%%%%%%%%%%%%%%%%%%%%%%%%%%%%%%%%%%%%%%%%%%%%%%%%%%%%%%%%
%%%%%%%%%%%%%%%%%%%%%%%%%%%%%%%%%%%%%%%%%%%%%%%%%%%%%%%%%%%%%%%%%%%%%%%%%%%%%%%%%%%%%%%%%%%%%%%%%%%%%%%%%
%%%%%%%%%%%%%%%%%%%%%%%%%%%%%%%%%%%%%%%%%%%%%%%%%%%%%%%%%%%%%%%%%%%%%%%%%%%%%%%%%%%%%%%%%%%%%%%%%%%%%%%%%
%%%%%%%%%%%%%%%%%%%%%%%%%%%%%%%%%%%%%%%%%%%%%%%%%%%%%%%%%%%%%%%%%%%%%%%%%%%%%%%%%%%%%%%%%%%%%%%%%%%%%%%%%
%%%%%%%%%%%%%%%%%%%%%%%%%%%%%%%%%%%%%%%%%%%%%%%%%%%%%%%%%%%%%%%%%%%%%%%%%%%%%%%%%%%%%%%%%%%%%%%%%%%%%%%%%
%%%%%%%%%%%%%%%%%%%%%%%%%%%%%%%%%%%%%%%%%%%%%%%%%%%%%%%%%%%%%%%%%%%%%%%%%%%%%%%%%%%%%%%%%%%%%%%%%%%%%%%%%
%%%%%%%%%%%%%%%%%%%%%%%%%%%%%%%%%%%%%%%%%%%%%%%%%%%%%%%%%%%%%%%%%%%%%%%%%%%%%%%%%%%%%%%%%%%%%%%%%%%%%%%%%
\section{Introduction}
%%%%%%%%%%%%%%%%%%%%%%%%%%%%%%%%%%%%%%%%%%%%%%%%%%%%%%%%%%%%%%%%%%%%%%%%%%%%%%%%%%%%%%%%%%%%%%%%%%%%%%%%%
%%%%%%%%%%%%%%%%%%%%%%%%%%%%%%%%%%%%%%%%%%%%%%%%%%%%%%%%%%%%%%%%%%%%%%%%%%%%%%%%%%%%%%%%%%%%%%%%%%%%%%%%%
%%%%%%%%%%%%%%%%%%%%%%%%%%%%%%%%%%%%%%%%%%%%%%%%%%%%%%%%%%%%%%%%%%%%%%%%%%%%%%%%%%%%%%%%%%%%%%%%%%%%%%%%%
%%%%%%%%%%%%%%%%%%%%%%%%%%%%%%%%%%%%%%%%%%%%%%%%%%%%%%%%%%%%%%%%%%%%%%%%%%%%%%%%%%%%%%%%%%%%%%%%%%%%%%%%%
%%%%%%%%%%%%%%%%%%%%%%%%%%%%%%%%%%%%%%%%%%%%%%%%%%%%%%%%%%%%%%%%%%%%%%%%%%%%%%%%%%%%%%%%%%%%%%%%%%%%%%%%%

Stochastic gradient descent (SGD) and its variants, including AdaGrad
\cite{Duchi}, RMSProp \cite{Tieleman2012} and Adam
\cite{Kingma2015}, are the main workhorses of modern machine
learning and deep learning.
These methods are quite sensitive to tuning the learning
rate, which usually requires testing different alternatives and
evaluating them on a validation data set.  When possible, this adds
significantly to the computational cost of using them.  However,
there are also situations where proper validation is difficult, such
as differentially private or federated learning.  In these settings,
effective adaptive algorithms can be extremely important for efficient
learning.
While adaptive SGD alternatives such as AdaGrad, RMSProp and Adam are
not as sensitive to tuning as plain SGD, they nevertheless require
tuning for good performance.  Furthermore, in \cite{Wilson2017} it is
argued that commonly used adaptive methods such as AdaGrad, RMSProp
and Adam can lead to very poor generalisation performance in deep
learning and that properly tuned basic SGD is a very competitive
approach.

Differential privacy (DP) \cite{dwork_et_al_2006}
has recently risen as the dominant paradigm for privacy-preserving
machine learning. A number of differentially private algorithms have
been proposed addressing both important specific models
(e.g.~\cite{chaudhuri08privacy}; \cite{Dwork2014}; \cite{Abadi2016})
as well as more general
approaches to learning
(e.g.~\cite{Chaudhuri2011}; \cite{Dimitrakakis2014}; \cite{zhang16differential}; \cite{Park2016}; \cite{Jalko2017}).
Like in machine learning more generally, differentially private
stochastic gradient descent (DP-SGD)
\cite{Rajkumar2012,Song2013,Abadi2016} has emerged as an important
tool for implementing differential privacy for a number of
applications. The introduction of very tight bounds on the privacy
loss occurring during the iterative algorithm computed via the moments
accountant \cite{Abadi2016} has made these algorithms particularly
attractive. Furthermore, DP's invariance to post-processing means that
the same privacy guarantees apply to any algorithm that uses the same
gradient information, including adaptive and accelerated methods.
In addition to deep learning, stochastic gradients and more recently
the moments accountant have been
used in algorithms for other paradigms, such as Bayesian inference
\cite{wang15privacy,Jalko2017,Li2017}.

It is clear that standard hyperparameter tuning methods typically used
for tuning the learning rate are not directly applicable to DP
learning because of the need to account for the additional privacy
loss for multiple runs of the learning and validation set use.  Most
previous work glosses this over, with two notable exceptions.
Kusner et al. \cite{Kusner2015} presented DP Bayesian optimisation that accounts
for the privacy loss for the validation set, but they completely
ignore training set privacy.  Very recently Liu and Talwar \cite{Liu2018}
introduced DP meta selection for DP hyperparameter tuning, but their
approach imposes a 2-3x loss in privacy and only supports random
hyperparameter search which may carry significant computational cost.

Federated learning \cite{mcmahan2017} has become popular as a means
for communication-efficient learning with distributed data, and a
useful tool for further improving privacy as well. The
federated setup can severely restrict the possibility of using
validation, because individual clients may differ from each other
significantly which limits the value of generalising hyperparameters
across clients, while at the same time limited availability of data
and compute at an individual client may limit the use of local
validation. In an extreme case the distribution of samples may be
extremely biased between different clients, requiring the use of very
different local learning rates that are impossible to tune with
classical methods. Adaptive learning rate tuning can greatly increase
both learning efficiency and stability in such cases.

In this paper, we propose a rigorous adaptive method for finding a
good learning rate for SGD, and apply in DP and federated learning
settings. The adaptation is performed during
learning, which implies that the learning process only has to be
executed once, leading to savings in compute time and efficient use of
the privacy budget. We prove the privacy of our method based on the
moments accountant mechanism.

\subsection{Main contributions}

We propose the first learning rate adaptive DP SGD
method. We give rigorous moment bounds
for the method, and using these bounds, we can compute tight $(\varepsilon,\delta)$-bounds
using the so called moments accountant technique. By simple derivations, we show
how to determine the additional tolerance hyperparameter in the algorithm.
In computational experiments we show that it is competitive with optimally
tuned standard optimisation methods without any tuning.
We further demonstrate that the method can help stabilise federated
learning especially when the data are non-uniformly distributed to
different clients.

%%%%%%%%%%%%%%%%%%%%%%%%%%%%%%%%%%%%%%%%%%%%%%%%%%%%%%%%%%%%%%%%%%%%%%%%%%%%%%%%%%%%%%%%%%%%%%%%%%%%%%%%%
%%%%%%%%%%%%%%%%%%%%%%%%%%%%%%%%%%%%%%%%%%%%%%%%%%%%%%%%%%%%%%%%%%%%%%%%%%%%%%%%%%%%%%%%%%%%%%%%%%%%%%%%%
%%%%%%%%%%%%%%%%%%%%%%%%%%%%%%%%%%%%%%%%%%%%%%%%%%%%%%%%%%%%%%%%%%%%%%%%%%%%%%%%%%%%%%%%%%%%%%%%%%%%%%%%%
%%%%%%%%%%%%%%%%%%%%%%%%%%%%%%%%%%%%%%%%%%%%%%%%%%%%%%%%%%%%%%%%%%%%%%%%%%%%%%%%%%%%%%%%%%%%%%%%%%%%%%%%%
%%%%%%%%%%%%%%%%%%%%%%%%%%%%%%%%%%%%%%%%%%%%%%%%%%%%%%%%%%%%%%%%%%%%%%%%%%%%%%%%%%%%%%%%%%%%%%%%%%%%%%%%%
\section{Motivation for the learning rate adaptation: extrapolation of differential equations}
%%%%%%%%%%%%%%%%%%%%%%%%%%%%%%%%%%%%%%%%%%%%%%%%%%%%%%%%%%%%%%%%%%%%%%%%%%%%%%%%%%%%%%%%%%%%%%%%%%%%%%%%%
%%%%%%%%%%%%%%%%%%%%%%%%%%%%%%%%%%%%%%%%%%%%%%%%%%%%%%%%%%%%%%%%%%%%%%%%%%%%%%%%%%%%%%%%%%%%%%%%%%%%%%%%%
%%%%%%%%%%%%%%%%%%%%%%%%%%%%%%%%%%%%%%%%%%%%%%%%%%%%%%%%%%%%%%%%%%%%%%%%%%%%%%%%%%%%%%%%%%%%%%%%%%%%%%%%%
%%%%%%%%%%%%%%%%%%%%%%%%%%%%%%%%%%%%%%%%%%%%%%%%%%%%%%%%%%%%%%%%%%%%%%%%%%%%%%%%%%%%%%%%%%%%%%%%%%%%%%%%%

%Our goal is to find a method which adapts the learning rate $\eta_\ell$ for Algorithm~\ref{alg:dpsgd}.
The main ingredient of the learning rate adaptation comes from numerical extrapolation of ordinary differential equations (ODEs), see e.g. \cite{hairer}.
We next describe this idea.
Let $g$ be a differentiable function $g \, : \, \mathbb{R}^d \rightarrow \mathbb{R}.$
Gradient descent (GD)
\begin{equation} \label{eq:GD}
\theta_{\ell+1} = \theta_\ell - \eta_\ell \nabla g(\theta_\ell)
\end{equation}
is a first-order method for finding a (local) minimum of the function $g$.
It can be seen as an explicit Euler method with a step of size $\eta_\ell$ applied to the system of ODEs
$\theta'(t) = - \nabla g (\theta)$, $\theta(0) = \theta_0 \in \mathbb{R}^d$,
% \begin{equation*} %\label{eq:ODE1}
% \frac{d}{dt} \theta(t) = - \nabla g (\theta), \quad \theta(0) = \theta_0 \in \mathbb{R}^d,
% \end{equation*}
which is also called the gradient flow corresponding to $g$.
%For basics of numerical methods for ODEs, we refer to~\cite{hairer}.

To get an estimate of the error made in the numerical approximation \eqref{eq:GD}, 
we extrapolate it as follows.
Consider one Euler step of size $\eta$ applied to the gradient flow,
\begin{equation}  \label{eq:x1}
\theta_1 = \theta_0- \eta \nabla g(\theta_0),
\end{equation}
and $\widehat{\theta}_1$ which is a result of two steps of size $\frac{\eta}{2}$:
\begin{equation*} 
	\begin{aligned}
\theta_{1/2} =  \theta_0 - \frac{\eta}{2} \nabla g(\theta_0), \quad 
\widehat{\theta}_1 =  \theta_{1/2} - \frac{\eta}{2} \nabla g(\theta_{1/2}).
	\end{aligned}
\end{equation*}
Using the Taylor expansion of the true solution $\theta(\eta)$, we see that
$\theta(\eta) - \theta_1 = \tfrac{\eta^2}{2}J_g(\theta_0) \nabla g(\theta_0) + \mathcal{O}(\eta^3)$
and that $2(\widehat{\theta}_1 - \theta_1)$ 
gives an $\mathcal{O}(\eta^3)$-estimate of the local error generated by the GD
step \eqref{eq:GD}. If at step $\ell$ of the iteration we have the error estimate
\begin{equation} \label{eq:err_1}
err_\ell = \norm{\widehat{\theta}_{\ell+1} - \theta_{\ell+1}},
\end{equation}
and if a local error of size $tol$ is desired, a simple mechanism for updating the step size is given by %~\cite[Ch.\;II]{hairer}
\begin{equation*} % \label{eq:control}
\eta_{\ell+1} = \min \bigg( \max \bigg( \frac{\mathrm{tol}}{ \mathrm{err}_\ell  }, \alpha_{\min} \bigg) ,\alpha_{\max} \bigg) \cdot \eta_\ell,
\end{equation*}
where $\alpha_{\min} < 1$ and $\alpha_{\max}>1$.
In our experiments we have used $\alpha_{\min} = 0.9, \alpha_{\max} = 1.1$.
In case $\nabla g$ has a large Lipschitz constant, a condition $\mathrm{err}_i < \mathrm{\mathrm{tol}}$
for the update $\theta_{\ell+1} \leftarrow \theta_\ell$ gives a more stable algorithm. This
procedure is depicted in Algorithm \ref{alg:adap}.

\begin{algorithm}
\caption{The update mechanism defined by the parameters $\alpha_{\min}$, $\alpha_{\max}$, $\mathrm{tol}$ }
\begin{algorithmic}
\STATE{Evaluate: $\mathrm{err}_\ell \leftarrow  \norm{\widehat{\theta}_{\ell+1} - \theta_{\ell+1}}$ }
\IF{$\mathrm{err}_\ell > \mathrm{\mathrm{tol}}$:}
	\STATE{$\theta_{\ell+1} \leftarrow \theta_\ell$ \quad (discard)}
\ENDIF
\STATE{update:
$\eta_{\ell+1} = \min ( \max (\frac{\mathrm{tol}}{ \mathrm{err}_\ell  }, \alpha_{\min} ) ,\alpha_{\max}) \cdot \eta_\ell$. }
\end{algorithmic}
\label{alg:adap}
\end{algorithm}

We apply Algorithm \ref{alg:adap} to the differentially private SGD method and to the federated learning algorithm.
The challenge in the DP setting is that for privacy reasons 
%we only have access to the outputs of the
%mechanism $\mathcal{M}$, so 
the gradients are blurred by the additive DP-noise.

%both by the noise arising from the minibatch approximations and by the additive DP-noise.

%%%%%%%%%%%%%%%%%%%%%%%%%%%%%%%%%%%%%%%%%%%%%%%%%%%%%%%%%%%%%%%%%%%%%%%%%%%%%%%%%%%%%%%%%%%%%%%%%%%%%%%%%
%%%%%%%%%%%%%%%%%%%%%%%%%%%%%%%%%%%%%%%%%%%%%%%%%%%%%%%%%%%%%%%%%%%%%%%%%%%%%%%%%%%%%%%%%%%%%%%%%%%%%%%%%
%%%%%%%%%%%%%%%%%%%%%%%%%%%%%%%%%%%%%%%%%%%%%%%%%%%%%%%%%%%%%%%%%%%%%%%%%%%%%%%%%%%%%%%%%%%%%%%%%%%%%%%%%
%%%%%%%%%%%%%%%%%%%%%%%%%%%%%%%%%%%%%%%%%%%%%%%%%%%%%%%%%%%%%%%%%%%%%%%%%%%%%%%%%%%%%%%%%%%%%%%%%%%%%%%%%
%%%%%%%%%%%%%%%%%%%%%%%%%%%%%%%%%%%%%%%%%%%%%%%%%%%%%%%%%%%%%%%%%%%%%%%%%%%%%%%%%%%%%%%%%%%%%%%%%%%%%%%%%
\section{Differential Privacy}
%%%%%%%%%%%%%%%%%%%%%%%%%%%%%%%%%%%%%%%%%%%%%%%%%%%%%%%%%%%%%%%%%%%%%%%%%%%%%%%%%%%%%%%%%%%%%%%%%%%%%%%%%
%%%%%%%%%%%%%%%%%%%%%%%%%%%%%%%%%%%%%%%%%%%%%%%%%%%%%%%%%%%%%%%%%%%%%%%%%%%%%%%%%%%%%%%%%%%%%%%%%%%%%%%%%
%%%%%%%%%%%%%%%%%%%%%%%%%%%%%%%%%%%%%%%%%%%%%%%%%%%%%%%%%%%%%%%%%%%%%%%%%%%%%%%%%%%%%%%%%%%%%%%%%%%%%%%%%
%%%%%%%%%%%%%%%%%%%%%%%%%%%%%%%%%%%%%%%%%%%%%%%%%%%%%%%%%%%%%%%%%%%%%%%%%%%%%%%%%%%%%%%%%%%%%%%%%%%%%%%%%

%%%%%%%%%%%%%%%%%%%%%%%%%%%%%%%%%%%%%%%%%%%%%%%%%%%%%%%%%%%%%%%%%%%%%%%%%%%%%%%%%%%%%%%%%%%%%%%%%%%%%%%%%
%%%%%%%%%%%%%%%%%%%%%%%%%%%%%%%%%%%%%%%%%%%%%%%%%%%%%%%%%%%%%%%%%%%%%%%%%%%%%%%%%%%%%%%%%%%%%%%%%%%%%%%%%
\subsection{Definition of Differential Privacy}
%%%%%%%%%%%%%%%%%%%%%%%%%%%%%%%%%%%%%%%%%%%%%%%%%%%%%%%%%%%%%%%%%%%%%%%%%%%%%%%%%%%%%%%%%%%%%%%%%%%%%%%%%
%%%%%%%%%%%%%%%%%%%%%%%%%%%%%%%%%%%%%%%%%%%%%%%%%%%%%%%%%%%%%%%%%%%%%%%%%%%%%%%%%%%%%%%%%%%%%%%%%%%%%%%%%

%
% Let the space of data be denoted by $\mathcal{X}$ and points $X,Y \in \mathcal{X}^n$.
% Let $d(X,Y)$ be the Hamming distance between $X$ and $Y$. E.g. if one data point difference,
% $d(X,Y) = 1$...

We first recall some basic definitions of differential privacy~\cite{DworkRoth}.
We use the following notation. An input set containing $N$ data points is denoted as
$X = (x_1,\ldots,x_N) \in \mathcal{X}^N$, where $x_i \in \mathcal{X}$, $1 \leq i \leq N $.
For giving the definition of the actual differential privacy we need the following definition.

\begin{dfn}
We say that two data sets $X$ and $X'$ are adjacent if they only differ in one record.
i.e., if $x_i \neq x_i'$ for some $i$, where $x_i \in X$ and $x_i' \in X'$.
\end{dfn}

The following definition formalises the $(\varepsilon,\delta)$-differential privacy of a randomised mechanism $\mathcal{M}$.

\begin{dfn}
Let $\varepsilon > 0$ and $\delta \in [0,1]$. Mechanism $\mathcal{M} \, : \, \mathcal{X}^N \rightarrow Z$ 
is $(\varepsilon,\delta)$-DP 
if for every pair of neighbouring data sets $X$, $X'$ 
%if for every 
%pair $x \simeq_X x'$ 
and every measurable set $E \subset Z$ we have
$$
\mathrm{Pr}( \mathcal{M}(X) \in E ) \leq \ee^\varepsilon \mathrm{Pr} (\mathcal{M}(X') \in E ) + \delta.
$$
\end{dfn}

This definition is closed under post-processing which means that if a mechanism $\mathcal{A}$
is $(\varepsilon,\delta)$-differential private, then so is the mechanism $\mathcal{B} \circ \mathcal{A}$
for all functions $\mathcal{B}$ that do not depend on the data.

%Differential privacy places constraints on the difference between the outputs of two adjacent inputs $X$ and $X'$ by a random mechanism. 
Assuming $X$ and $X'$ differ only by one record $x_i$, then by observing the outputs, 
the ability of an attacker to tell whether the output has resulted from $X$ or $X'$ remains bounded.
Thus, the record $x_i$ is protected. 
As the record in which the two data sets differ is arbitrary, by definition, the protection applies for the whole data set.

%%%%%%%%%%%%%%%%%%%%%%%%%%%%%%%%%%%%%%%%%%%%%%%%%%%%%%%%%%%%%%%%%%%%%%%%%%%%%%%%%%%%%%%%%%%%%%%%%%%%%%%%%
%%%%%%%%%%%%%%%%%%%%%%%%%%%%%%%%%%%%%%%%%%%%%%%%%%%%%%%%%%%%%%%%%%%%%%%%%%%%%%%%%%%%%%%%%%%%%%%%%%%%%%%%%
\subsection{Moments accountant}
%%%%%%%%%%%%%%%%%%%%%%%%%%%%%%%%%%%%%%%%%%%%%%%%%%%%%%%%%%%%%%%%%%%%%%%%%%%%%%%%%%%%%%%%%%%%%%%%%%%%%%%%%
%%%%%%%%%%%%%%%%%%%%%%%%%%%%%%%%%%%%%%%%%%%%%%%%%%%%%%%%%%%%%%%%%%%%%%%%%%%%%%%%%%%%%%%%%%%%%%%%%%%%%%%%%

We next recall  some basic definitions and results concerning the moments accountant technique
which is an important ingredient for our proposed method and crucial for obtaining tight $(\varepsilon,\delta)$-privacy bounds
for the differentially private stochastic gradient descent. We refer to~\cite{Abadi2016} for more details.

\begin{dfn}
Let $\mathcal{M} \, : \, \mathcal{X}^N \rightarrow \mathcal{Y}$ be a randomised mechanism, and let 
$X$ and $X'$ be a pair of adjacent data sets. Let $\mathrm{aux}$ denote any auxiliary input that does
not depend on $X$ or $X'$. For an outcome $o \in \mathcal{Y}$, the privacy loss at $o$ is defined as
$$
c(o; \mathcal{M}, \mathrm{aux}, X, X') = \log 
\frac{  \mathrm{Pr}(\mathcal{M}(\mathrm{aux},X) = o)  }{ \mathrm{Pr}(\mathcal{M}(\mathrm{aux},X') = o) }.
$$

\end{dfn}

\begin{dfn}
$\lambda$$th$ moment generating function $\alpha_\mathcal{M}(\lambda; \mathrm{aux},X,X')$ is defined
as
\begin{equation*}
	\begin{aligned}
		& \alpha_\mathcal{M}(\lambda; \mathrm{aux},X,X')  =  \\ 
		& \quad \log \mathbb{E}_{o \sim \mathcal{M}(\mathrm{aux},X)} 
		\left(  \exp(\lambda c(o; \mathcal{M}, \mathrm{aux}, X, X')  )  \right).
	\end{aligned}
\end{equation*}
\end{dfn}

\begin{dfn}
Let $\mathcal{M} \, : \, \mathcal{X}^N \rightarrow \mathcal{Y}$ be a randomised mechanism, and let 
$X$ and $X'$ be a pair of adjacent data sets. Let $\mathrm{aux}$ denote any auxiliary input that does
not depend on $X$ or $X'$. The moments accountant with an integer parameter $\lambda$ is defined as
$$
\alpha_\mathcal{M}(\lambda) = \max\limits_{\mathrm{aux},X,X'}  \alpha_\mathcal{M}(\lambda; \mathrm{aux},X,X').
$$
\end{dfn}
The privacy of our proposed method is based on the composability theorem (\cite[Thm.\;2]{Abadi2016}):
\begin{thm} \label{thm:comp}
Suppose that $\mathcal{M}$ consists of a sequence of adaptive mechanisms $\mathcal{M}_1, \ldots, \mathcal{M}_k$,
where $\mathcal{M}_i \, : \, \prod_{j=1}^{i-1} \mathcal{Y}_j \times \mathcal{X} \rightarrow \mathcal{Y}_i$, 
and $\mathcal{Y}_i$ is in the range of the $i$$th$ mechanism, i.e., $\mathcal{M} =  \mathcal{M}_k \circ \ldots \circ \mathcal{M}_1$.
Then, for any $\lambda$
\begin{equation} \label{eq:comp}
	\alpha_\mathcal{M}(\lambda)  \leq \sum_{i=1}^k \alpha_{\mathcal{M}_i}(\lambda),
\end{equation}
where the auxiliary input for $\alpha_{\mathcal{M}_i}(\lambda)$ is defined as all $\alpha_{\mathcal{M}_j}(\lambda)$'s
outputs for $j<i$, and $\alpha_\mathcal{M}(\lambda)$ takes $\mathcal{M}_i$'s output, for $i<k$, as the auxiliary input.

Moreover, for any $\varepsilon > 0$, the mechanism $\mathcal{M}$ is $(\varepsilon,\delta)$-differentially private
for
\begin{equation} \label{eq:min_d}
	\delta = \min\limits_{\lambda} \exp(\alpha_\mathcal{M} (\lambda) - \lambda \varepsilon).
\end{equation}
\end{thm}

The inequality \eqref{eq:comp} gives an
upper bound for the total moment $\alpha_\mathcal{M}(\lambda)$ of an iterative algorithm $\mathcal{M}$
if the moments $\alpha_{\mathcal{M}_i}(\lambda)$ of each iteration $i$ are known.
%Important for the additivity of the moments is that the noise of the mechanisms $\mathcal{M}_i$ is pairwise independent.
Using \eqref{eq:min_d}, the privacy parameters $\varepsilon$ and $\delta$ can be numerically computed 
from $\alpha_\mathcal{M}(\lambda)$-values.

%%%%%%%%%%%%%%%%%%%%%%%%%%%%%%%%%%%%%%%%%%%%%%%%%%%%%%%%%%%%%%%%%%%%%%%%%%%%%%%%%%%%%%%%%%%%%%%%%%%%%%%%%
%%%%%%%%%%%%%%%%%%%%%%%%%%%%%%%%%%%%%%%%%%%%%%%%%%%%%%%%%%%%%%%%%%%%%%%%%%%%%%%%%%%%%%%%%%%%%%%%%%%%%%%%%
\subsection{Differentially private stochastic gradient descent}
%%%%%%%%%%%%%%%%%%%%%%%%%%%%%%%%%%%%%%%%%%%%%%%%%%%%%%%%%%%%%%%%%%%%%%%%%%%%%%%%%%%%%%%%%%%%%%%%%%%%%%%%%
%%%%%%%%%%%%%%%%%%%%%%%%%%%%%%%%%%%%%%%%%%%%%%%%%%%%%%%%%%%%%%%%%%%%%%%%%%%%%%%%%%%%%%%%%%%%%%%%%%%%%%%%%

Suppose we want to find a minimum (w.r.t. $\theta$) of a loss function %$\mathcal{L} \, : \, \mathbb{R}^d \times 
%\mathcal{X}^N \rightarrow \mathbb{R}$  
of the form
%\begin{equation*} %\label{eq:loss_fct}
	$\mathcal{L}(\theta,X) = \tfrac{1}{N} \sum_{i = 1}^N f(\theta,x_i)$.
%\end{equation*}
At each step of the differentially private SGD, %(see Algorithm~\ref{alg:dpsgd}), 
we compute the gradient $\nabla_\theta f(\theta,x_i)$ for a random minibatch $B$,
clip the 2-norm of each gradient belonging to the minibatch, compute the average, add noise in order to protect privacy, 
and take a GD step using this noisy gradient.
For a data set $X$, the basic mechanism is then given by
%\begin{equation} \label{eq:M}
	$\mathcal{M}(X) = \sum_{ i \in  B } \widetilde \nabla f(\theta,x_i) + \mathcal{N}(0,C^2 \sigma^2 I)$,
%\end{equation}
where $\widetilde \nabla f(\theta,x_i)$'s denote the gradients clipped with a constant $C>0$, i.e., 
$\norm{\widetilde \nabla f(\theta,x_i)} \leq C$ for all $i \in B$.
% We recall a result by \cite{Abadi2016}, which gives a privacy bound for the mechanism \eqref{eq:M}.
%
% \begin{lem}
% Suppose that $f \, : \, \mathcal{X} \rightarrow \mathbb{R}^d$ with $\norm{f( \cdot )} \leq 1$.
% Let $\sigma \geq 1$ and $B$ a minibatch with sampling probability $q$, i.e., $q = \tfrac{\abs{B}}{N}$, where $N$
% is the number of records in the data. If $q < \frac{1}{16 \sigma}$, then for any positive integer $\lambda \leq \sigma^2 \ln \frac{1}{q \sigma}$, the
% mechanism $\mathcal{M}(X) = \sum_{i \in B} f (\theta,x_i) + \mathcal{N}(0, \sigma^2 I)$ satisfies
% $$
% \alpha_\mathcal{M}(\lambda) \leq \frac{ q^2 \lambda(\lambda+1) }{(1-q) \sigma^2} + \mathcal{O} (q^3\lambda^3/\sigma^3).
% $$
% \end{lem}
In numerical experiments we compute the moments using the numerical methods of \cite{Abadi2016}.

\section{Adaptive DP algorithm}

The result of applying the learning rate adaptation to DP-SGD is depicted in Algorithm~\ref{alg:main}. We abbreviate this method as ADADP.
Instead of \eqref{eq:err_1}, we use for the error estimate the 2-norm of the function $err(\theta, \widehat{\theta})$, where
\begin{equation} \label{eq:err_fct}
	err(\theta,\widehat{\theta})_i = \frac{ | \theta_i - \widehat{\theta}_i |}{\max(1,\abs{\theta_i})}
\end{equation}
as this was found to perform better numerically. 
In the case of DP learning the algorithm was found to be stable
without the condition $\mathrm{err}_i < \mathrm{\mathrm{tol}}$ so we omit it.
Here the factor $\abs{B}$ is dropped, as it only scales the learning rate $\eta$.

\begin{algorithm}
\caption{ADADP update mechanism defined by the parameters $\alpha_{\min}$, $\alpha_{\max}$, $\mathrm{tol}$ }
\begin{algorithmic}	
\STATE{Draw a batch $B_1$, with probability $q = \abs{B}/N$.}
\STATE{Clip the gradients (with constant $C$) and evaluate at $\theta_{\ell}$:  
$$ 
G_1 \leftarrow \sum_{i \in B_1} \widetilde{\nabla } f_{\theta_{\ell}} (x_i) + \mathcal{N}(0, C^2 \sigma^2 I).
$$}
\STATE{Take one Euler step of size $\eta_\ell$:
$$
\theta_{\ell+1}  \leftarrow\theta_\ell - \eta_\ell G_1,
$$}	

 \STATE{Take a step of size $\frac{\eta_\ell}{2}$:
 \begin{equation*}
 	\begin{aligned}
 \theta_{\ell+1/2} \leftarrow  \theta_\ell - \frac{\eta_\ell}{2} G_1
 	\end{aligned}
 \end{equation*}}
\STATE{Draw a batch $B_2$, with probability $q = \abs{B}/N$,
clip the gradients (with constant $C$) and evaluate at $\theta_{\ell+1/2}$: 
$$ 
G_2 \leftarrow \sum_{i \in B_2} \widetilde{\nabla } f_{\theta_{\ell+1/2}} (x_i) + \mathcal{N}(0,C^2 \sigma^2 I).
$$}
 \STATE{Take a step of size $\frac{\eta_\ell}{2}$:
 \begin{equation*}
 	\begin{aligned}
 \widehat{\theta}_{\ell+1} \leftarrow  \theta_{\ell+1/2} - \frac{\eta_\ell}{2} G_2
 	\end{aligned}
 \end{equation*}}
\STATE{Evaluate: $\mathrm{err}_\ell \leftarrow  \norm{err(\theta_{\ell+1}, \widehat{\theta}_{\ell+1})}_2$ }
\STATE{Update: 
$\eta_{\ell+1} \leftarrow \min \big( \max \big(\frac{\mathrm{tol}}{ \mathrm{err}_\ell }, \alpha_{\min} \big) ,\alpha_{\max} \big) \cdot \eta_\ell.$ }
\end{algorithmic}
\label{alg:main}
\end{algorithm}

\subsection{Privacy preserving properties of the method}

By the very construction of Algorithm~\ref{alg:main} and due to the post-processing property of differential
privacy, we have the following result.

\begin{thm} \label{thm:main}
Let $q=\abs{B}/N$, $\sigma \geq 1$ and $C>0$. Let $\alpha_\mathcal{M}(\lambda)$ be the moments accountant
of a mechanism $\mathcal{M}$ for these parameter values. Let $\widetilde{\mathcal{M}}$ denote
the mechanism of Algorithm~\ref{alg:main} using these parameter values. Then,
$$
\alpha_{\widetilde{\mathcal{M}}}(\lambda) \leq 2 \alpha_\mathcal{M}(\lambda).
$$
\end{thm}
By Theorem~\ref{thm:main}, using the same parameter values, we are allowed to run Algorithm~\ref{alg:main}
half as many times as differentially private SGD in order to have the same privacy.

\subsection{Choice of parameter $\mathrm{tol}$} \label{Sec:noise}

For simplicity, consider the situation where we apply DP-SGD
with step sizes $\{\eta_\ell \}$. After $T$ steps
\begin{equation} \label{eq:t_T}
	\theta_T = \theta_0 - \sum_{\ell=0}^{T-1} \eta_\ell g(\theta_\ell) + 
	\mathcal{N}\left(0,\sum_{\ell=0}^{T-1} \eta_\ell^2 C^2 \sigma^2 I \right),
\end{equation}
where 
$g(\theta_\ell) = \sum_{i \in B_\ell} \widetilde{\nabla } f (x_i)$.
Clearly, $\norm{g(\theta_\ell)}_2 \leq C \abs{B}$. 
It holds %$\theta_{\ell+1} - \widehat{\theta}_{\ell+1}$ it holds
\begin{equation*}% \label{eq:local_est}
    \norm{\theta_{\ell+1} - \widehat{\theta}_{\ell+1}} = \frac{\eta_\ell}{2}
    \norm{ g(\theta_\ell) - g(\widehat{\theta}_\ell)  + \mathcal{N}(0,2 C^2 \sigma^2 I) }.
\end{equation*}
Assuming $\sqrt{d} \gg \abs{B}$ (Recall: $\theta \in \mathbb{R}^d$) 
and taking the expectation value, we may approximate
$\norm{\theta_{\ell+1} - \widehat{\theta}_{\ell+1}}  \approx 
\eta_\ell \sigma C \sqrt{ \tfrac{d}{2}  }.$
If we set this estimate to $\mathrm{tol}$, we have approximately
$\eta_\ell^2 = \tfrac{2 \mathrm{tol}^2}{\sigma^2 C^2 d}$.
Substituting this into the third term on the right hand side of \eqref{eq:t_T}, we see that
after $T$ steps each element of that term is approximately $\tfrac{\sqrt{2T} \mathrm{tol}}{\sqrt{d}}.$
By requiring that this noise is, for example, $\mathcal{O}(1)$, we find 
a suitable value for the parameter $\mathrm{tol}$. In our experiments with neural networks
$\tfrac{2T}{d} = \mathcal{O}(1)$ and we use $\mathrm{tol} = 1.0$. In the other extreme, i.e., $\sqrt{d} \ll \abs{B}$, 
the term $g(\theta_\ell) - g(\widehat{\theta}_\ell)$ is likely to dominate the estimate $\norm{\theta_{\ell+1} - \widehat{\theta}_{\ell+1}}$. % \eqref{eq:local_est}.
Then it is the Lipschitz constant of $g$ that dictates the suitable step size $\eta_\ell$ and potentially a smaller value 
of $\mathrm{tol}$ is needed.

%%%%%%%%%%%%%%%%%%%%%%%%%%%%%%%%%%%%%%%%%%%%%%%%%%%%%%%%%%%%%%%%%%%%%%%%%%%%%%%%%%%%%%%%%%%%%%%%%%%%
%%%%%%%%%%%%%%%%%%%%%%%%%%%%%%%%%%%%%%%%%%%%%%%%%%%%%%%%%%%%%%%%%%%%%%%%%%%%%%%%%%%%%%%%%%%%%%%%%%%%
\section{Adaptive federated avearaging algorithm}
%%%%%%%%%%%%%%%%%%%%%%%%%%%%%%%%%%%%%%%%%%%%%%%%%%%%%%%%%%%%%%%%%%%%%%%%%%%%%%%%%%%%%%
%%%%%%%%%%%%%%%%%%%%%%%%%%%%%%%%%%%%%%%%%%%%%%%%%%%%%%%%%%%%%%%%%%%%%%%%%%%%%%%%%%%%%%

We consider next the federated averaging algorithm given by \cite{mcmahan2017}. 
The idea is such that the same model is first distributed to several clients. % which all have their own separate data.
The clients update their models based on their local data, and these models are then aggregated after a given interval by a server which then averages the
models to obtain a global model. This global model is then again distributed to the clients. 

In the algorithm described in \cite[Algorithm 1]{mcmahan2017}, a
random subset of clients is considered at each aggregation. We
consider the case $C=1$ where each client participates in every
aggregation, and replace the gradient step in client update with a
non-private variant of Algorithm \ref{alg:main}.

In \cite{mcmahan2017}, SGD with a constant learning rate is used for the updates of the clients.
The motivation for using the learning rate adaptation comes from the fact that after averaging and distributing,
the model at each client may be very far from the optimum for the local data and thus small steps are needed in the beginning of each sub training.
Moreover, the data may vary considerably between the clients, leading to varying optimal learning rates.

For the learning rate adaptation, we use the same procedure as in ADADP, but without the additive noise and clipping of the gradients.
We also add the condition $\mathrm{err}_i < \mathrm{\mathrm{tol}}$ for the model update as it makes the algorithm
considerably more stable. This adaptive client update is equivalent to Algorithm~\ref{alg:main} without clipping and additive noise 
(i.e., $C=\infty$ and $C\sigma=0$). We use here also the error function \eqref{eq:err_fct}.
In all federated learning experiments, we used $\mathrm{tol}=0.1$ (see the discussion of Subsection~\ref{Sec:noise}).

%%%%%%%%%%%%%%%%%%%%%%%%%%%%%%%%%%%%%%%%%%%%%%%%%%%%%%%%%%%%%%%%%%%%%%%%%%%%%%%%%%%%%%%%%%%%%%%%%%%%
%%%%%%%%%%%%%%%%%%%%%%%%%%%%%%%%%%%%%%%%%%%%%%%%%%%%%%%%%%%%%%%%%%%%%%%%%%%%%%%%%%%%%%%%%%%%%%%%%%%%
\section{Experiments}
%%%%%%%%%%%%%%%%%%%%%%%%%%%%%%%%%%%%%%%%%%%%%%%%%%%%%%%%%%%%%%%%%%%%%%%%%%%%%%%%%%%%%%
%%%%%%%%%%%%%%%%%%%%%%%%%%%%%%%%%%%%%%%%%%%%%%%%%%%%%%%%%%%%%%%%%%%%%%%%%%%%%%%%%%%%%%

We compare ADADP with Adam combined with DP gradients. %~\cite{kingma2014adam}.
The federated averaging algorithm with adaptive learning rates is compared to constant learning rate SGD.
We compare the methods on two standard datasets: MNIST~\cite{Lecun} and CIFAR-10. %~\cite{cifar}.

The random sampling of minibatches is approximated as in \cite{Abadi2016}, i.e., by
randomly permuting the data elements and then partitioning them into minibatches of a fixed size.
As we see, Algorithm~\ref{alg:main} needs two minibatches per iteration:
one to compute the vector $G_1$ and then the next one to compute $G_2$. Therefore, in one epoch we run
$\frac{N}{2\abs{B}}$ iterations. Then the number of gradient evaluations per epoch is the same as for SGD and Adam
and thus the computation times are essentially equivalent.
 When using ADADP, also the per epoch privacy cost is then the same
for all the methods considered, for a fixed value of the noise parameter $\sigma$. 

In the DP setting the methods are compared by measuring the test accuracy for a given
$\varepsilon$-value, when $\delta = 10^{-5}$. The $\varepsilon$-values are computed using the moments accountant method
described in \cite{Abadi2016}.

The values $\alpha_{\min} = 0.9$ and $\alpha_{\max} = 1.1$ were used in all experiments. 
 In all experiments with DP we used the value $\mathrm{tol}=1.0$ 
and in all non-DP experiments the value $\mathrm{tol}=0.1$.

All experiments are implemented using PyTorch.

 %%%%%%%%%%%%%%%%%%%%%%%%%%%%%%%%%%%%%%%%%%%%%%%%%%%%%%%%%%%%%%%%%%%%%%%%%%%%%%%%%%%%%%
%%%%%%%%%%%%%%%%%%%%%%%%%%%%%%%%%%%%%%%%%%%%%%%%%%%%%%%%%%%%%%%%%%%%%%%%%%%%%%%%%%%%%%
 \subsection{Datasets and test architectures}
%%%%%%%%%%%%%%%%%%%%%%%%%%%%%%%%%%%%%%%%%%%%%%%%%%%%%%%%%%%%%%%%%%%%%%%%%%%%%%%%%%%%%%
%%%%%%%%%%%%%%%%%%%%%%%%%%%%%%%%%%%%%%%%%%%%%%%%%%%%%%%%%%%%%%%%%%%%%%%%%%%%%%%%%%%%%%

In MNIST each example is a $28 \times 28$ size gray-level image. The training set contains 60000 and the test set
10000 examples.
For MNIST we use a feedforward neural network with 2 hidden layers with
$256$ hidden units. As a result, the total number of parameters for this network is $334336$.
We use ReLU units and the last layer is passed to softmax of $10$ classes with
cross-entropy loss. Without additional noise ($\sigma=0$) we reach an accuracy of around $96\%$.

CIFAR-10 consists of colour images classified into 10 classes.
The training set contains 50000 and the test set 10000 examples.
Each example is a $32 \times 32$ image with three RGB channels.
The CIFAR-100 dataset has similar images classified into 100 classes.
For CIFAR-10 we use a simple neural network, which consists of two convolutional
layers followed by three fully connected layers. The convolutional layers use $3 \times 3$
convolutions with stride $1$, followed by ReLU and max pools, with 64 channels each. 
The output of the second convolutional layer is flattened into a vector of dimension $1600$.
The fully connected layers have $500$ hidden units. Last layer
is passed to softmax of $10$ classes with cross-entropy loss.
The total number of parameters for this network is about $10^6$. 
Similarly to the experiments of \cite{Abadi2016}, 
in the DP setting we pre-train the convolutional layers using the CIFAR-100 data set and the differentially private
optimisation is carried out only for the fully connected layers.

%%%%%%%%%%%%%%%%%%%%%%%%%%%%%%%%%%%%%%%%%%%%%%%%%%%%%%%%%%%%%%%%%%%%%%%%%%%%%%%%%%%%%%%%%%%%%%%%%%%%
%%%%%%%%%%%%%%%%%%%%%%%%%%%%%%%%%%%%%%%%%%%%%%%%%%%%%%%%%%%%%%%%%%%%%%%%%%%%%%%%%%%%%%%%%%%%%%%%%%%%
\subsection{Federated learning experiments}
%%%%%%%%%%%%%%%%%%%%%%%%%%%%%%%%%%%%%%%%%%%%%%%%%%%%%%%%%%%%%%%%%%%%%%%%%%%%%%%%%%%%%%
%%%%%%%%%%%%%%%%%%%%%%%%%%%%%%%%%%%%%%%%%%%%%%%%%%%%%%%%%%%%%%%%%%%%%%%%%%%%%%%%%%%%%%

\begin{figure}
\centering
\begin{subfigure}{0.9\textwidth}
  \centering
  \includegraphics[width=1.0\linewidth]{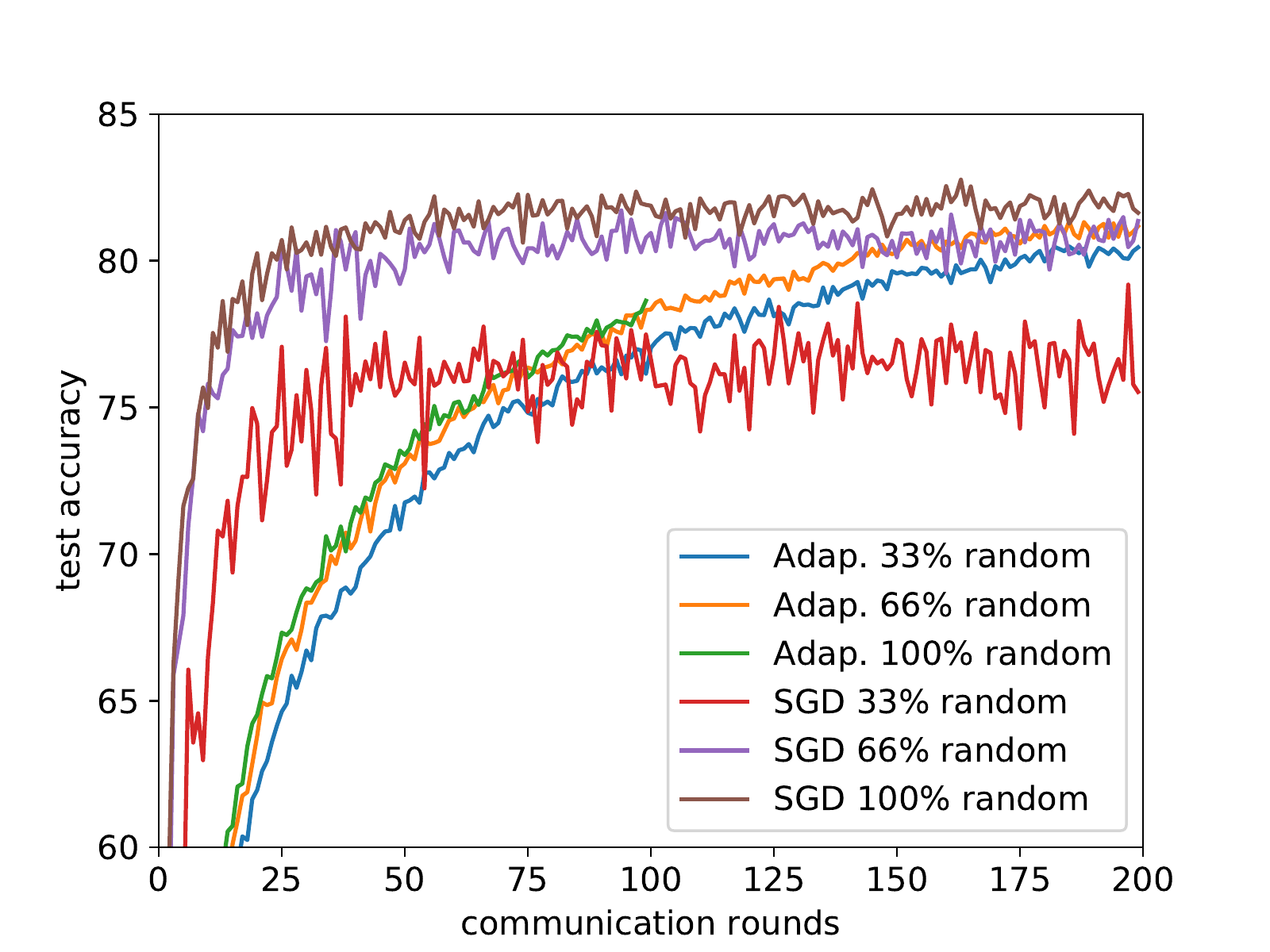}
  \caption{Interpolated case.}
  \label{fig:fed2}
\end{subfigure} \\
\begin{subfigure}{0.9\textwidth}
  \centering
  \includegraphics[width=1.0\linewidth]{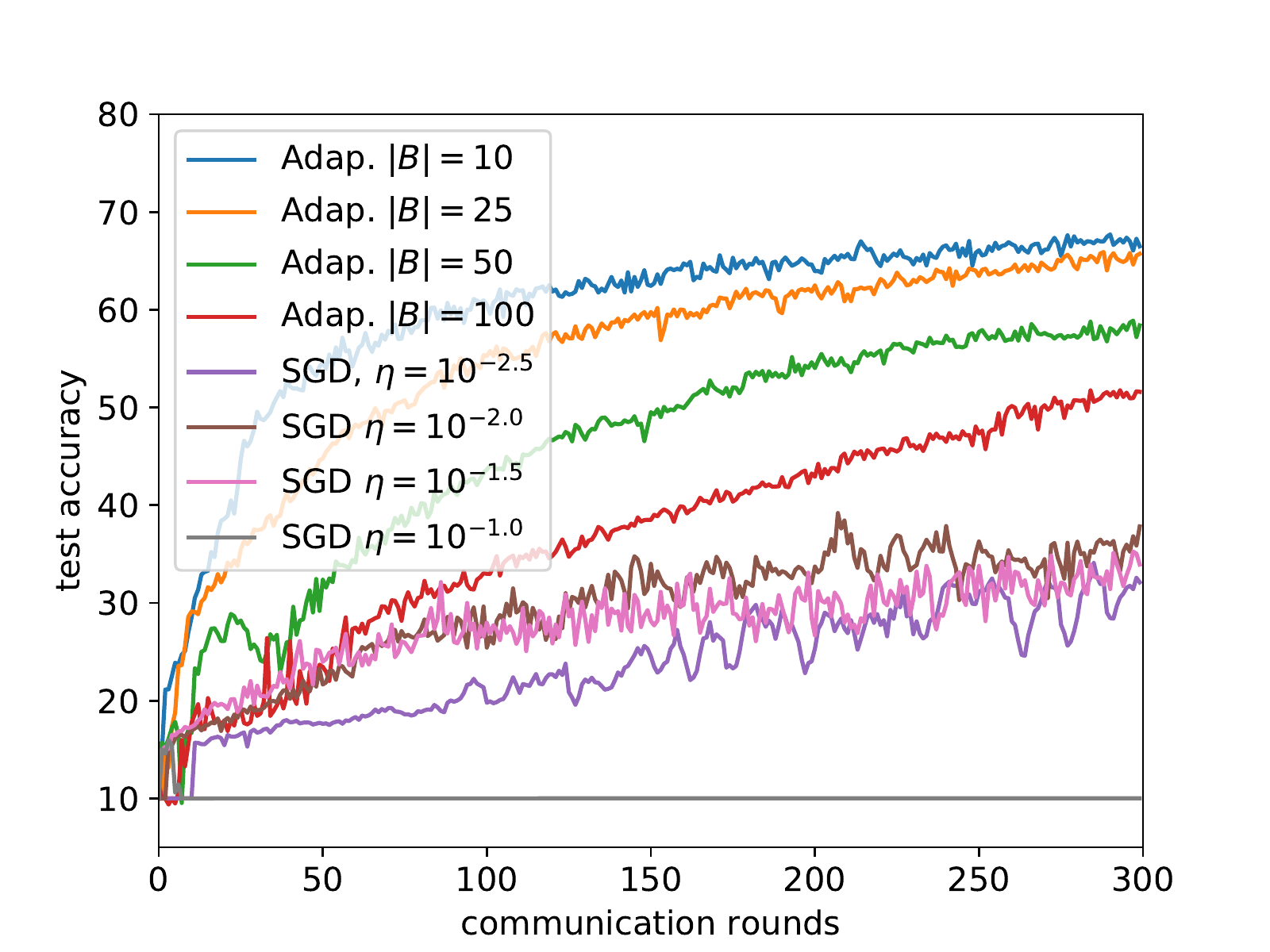}
  \caption{0\% random case.}
  \label{fig:fed1}
\end{subfigure}%
\caption{CIFAR-10 test accuracies for the federated averaging algorithm using the adaptive method and a learning rate tuned SGD, 
when the training data is interpolated
between the pathological case and the uniformly random distribution of data.}
\label{fig:test}
\end{figure}

We consider a pathological case, where the CIFAR-10 training data is divided to five clients 
such that client 1 has cars and trucks, client 2 planes and ships, client 3 cats and dogs, 
client 4 birds and frogs and client 5 deers and horses. Then, each client has 10000 images.
We interpolate between this pathological case and a uniformly random distribution of data between the five clients.
Figure~\ref{fig:fed2} depicts the test accuracies for the learning rate adaptive algorithm and SGD.
% when at each client a given percentage of the data comes from a uniformly random distribution. 
 The learning rate of SGD is tuned in the grid $\{\ldots, 10^{-2.5}, 10^{-2.0}, 10^{-1.5}, \ldots\}$.
 We use in all alternatives $\abs{B} = 10$.
We see that as the distribution of data becomes more pathological ($33\%$ of the data chosen randomly), the learning rate
adaptive method is able to maintain the overall performance much better than SGD.
Figure~\ref{fig:fed1} corresponds here to the fully pathological case.
For a given minibatch size $\abs{B}$,
each client carries out $E$ number of sub steps between each aggregation
such that $\abs{B} \cdot E = 10000$ (one epoch of data for each client).

\begin{figure}
\centering
\begin{subfigure}{0.9\textwidth}
  \centering
  \includegraphics[width=1.0\linewidth]{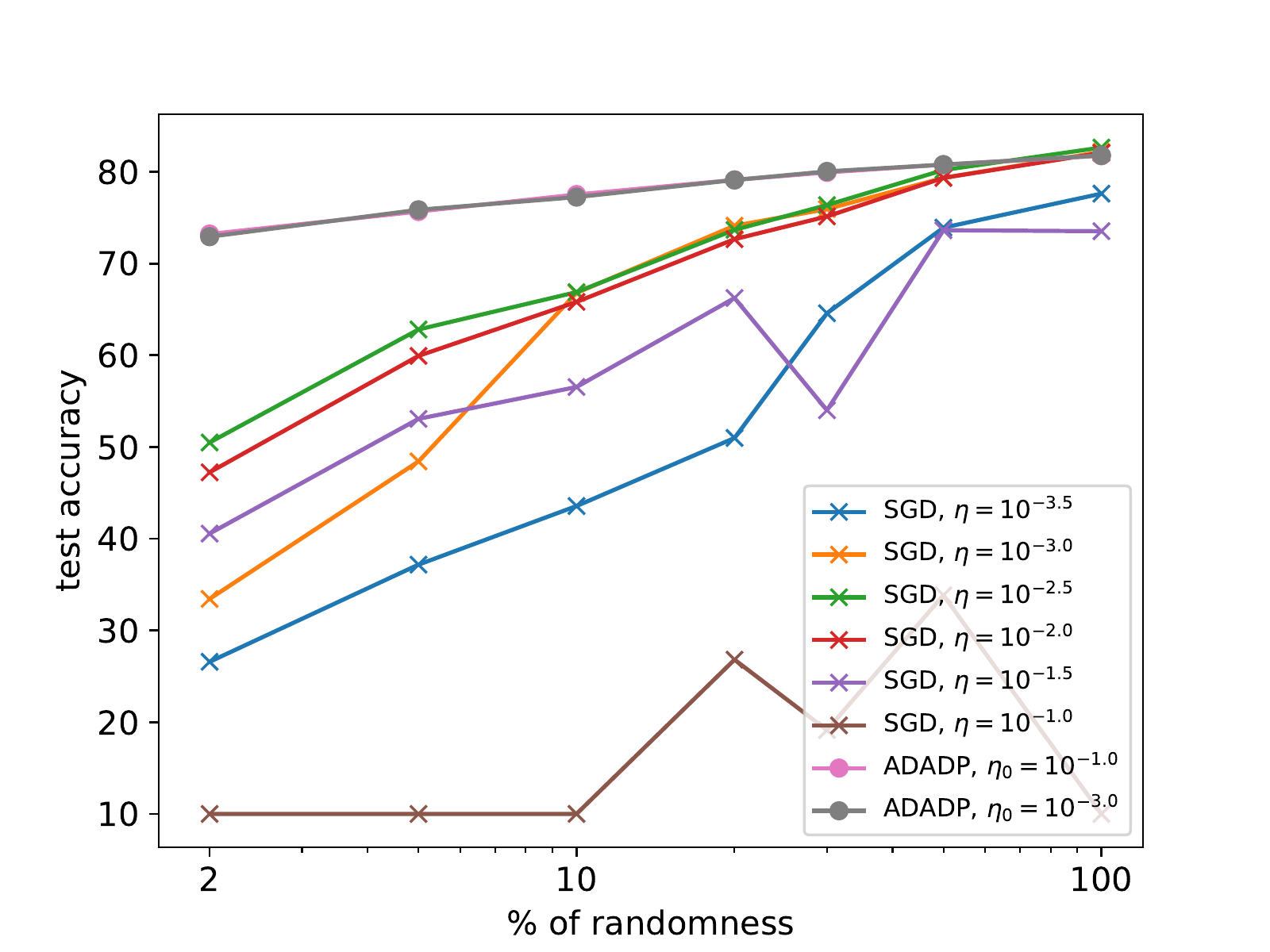}
  \caption{Test accuracy vs. \% of randomness.   }
  \label{fig:fed3}
\end{subfigure} \\
\begin{subfigure}{0.9\textwidth}
  \centering
  \includegraphics[width=1.0\linewidth]{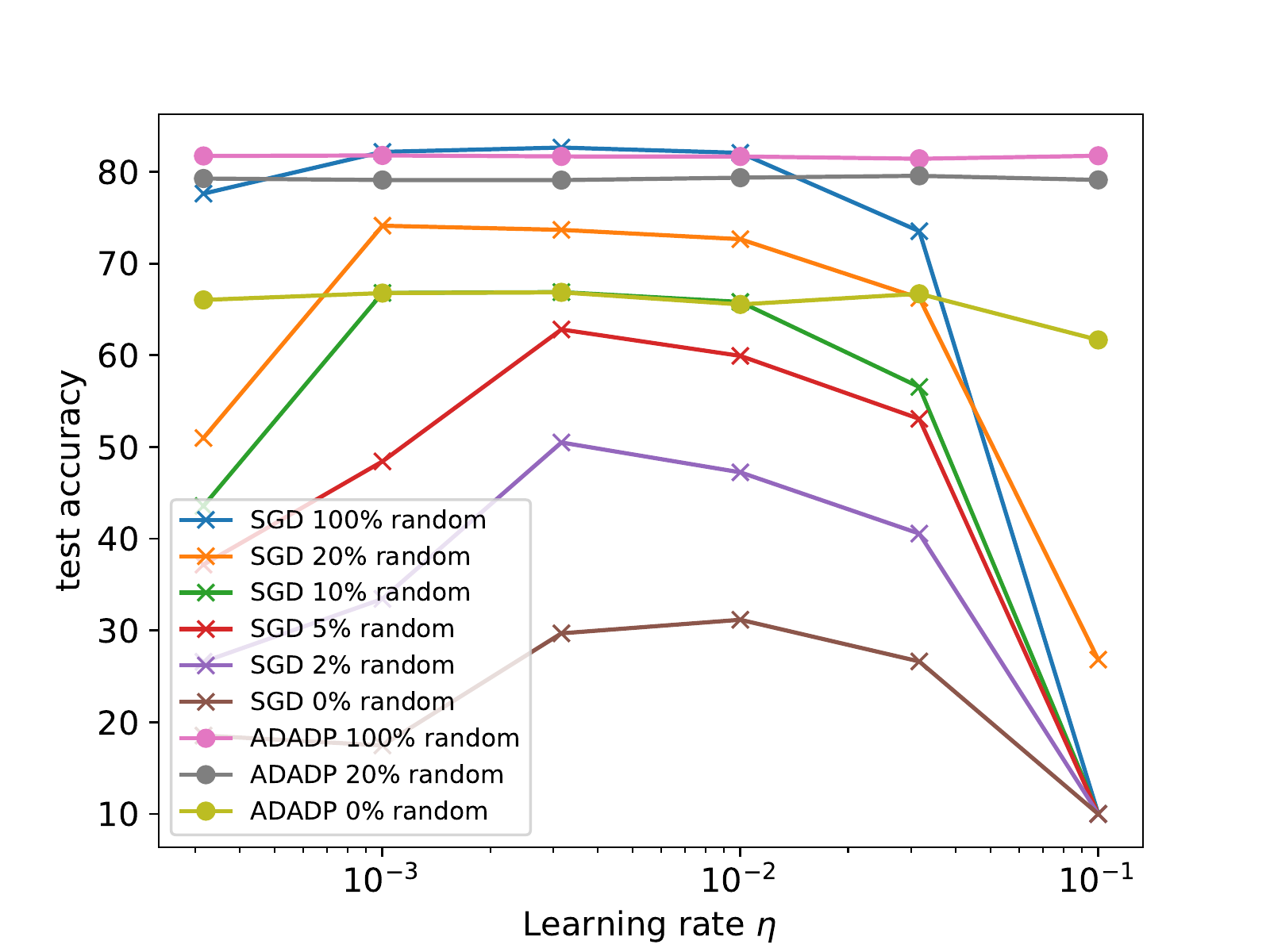}
  \caption{Test accuracy vs. learning rate $\eta$ for different levels of randomness. For ADADP $\eta$ means here 	the initial $\eta_0$.}
  \label{fig:fed4}
\end{subfigure}
\caption{CIFAR-10 test accuracies  for the federated averaging algorithm after 200 communication rounds using ADADP for different initial learning rates $\eta_0$ 
and constant learning rate SGD for different $\eta$.
The training data is interpolated between the pathological case and the uniformly random distribution of data.
All points are averages of three runs.}
\label{fig:test2}
\end{figure}

Figure \ref{fig:test2} illustrates further how ADADP is able to adapt even for highly pathological distribution of data
whereas the performance of (even an optimally tuned) SGD reduces drastically when the data becomes less uniformly distributed.

Adam gave poor results in this example. Figure~\ref{fig:fed3} shows the test accuracies in the
interpolated case, where $33\%$ of the data is chosen randomly for each client, for the best initial learning rates
found from the grid $\{\ldots, 10^{-5.5}, 10^{-5.0}, 10^{-4.5}, \ldots\}$. We use here $\abs{B} = 10$.
 Notice here the different scale of y-axis as in Figure~\ref{fig:fed2}.

\begin{figure}[tb]
\begin{center}
\includegraphics[scale=.68]{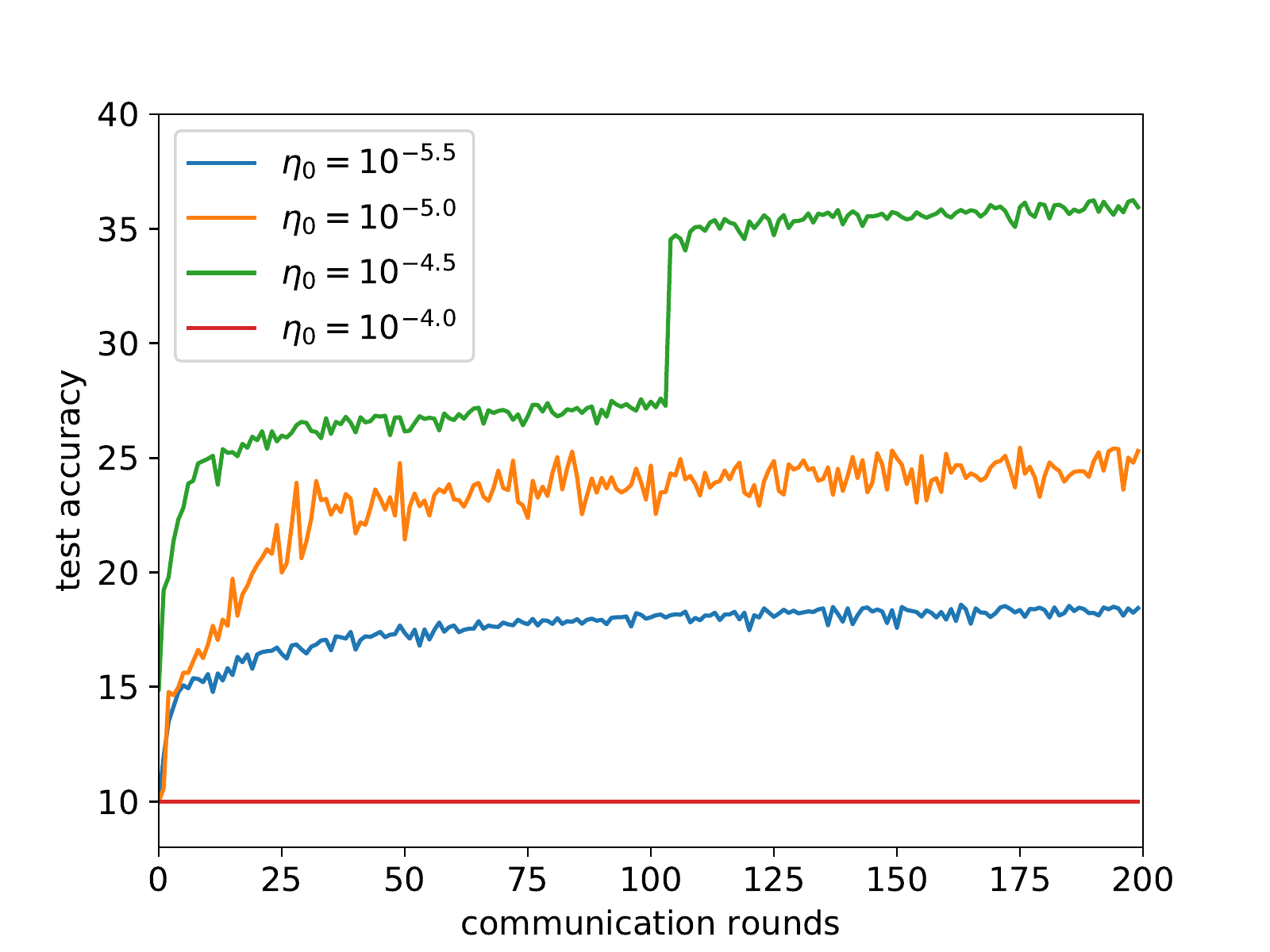}
\end{center}
\caption{Test accuracies for the federated averaging algorithm, when the client updates are done using Adam using different
initial learning rates $\eta_0$. This is the interpolated case, where $33\%$ of the data is chosen randomly for each client.}
\label{fig:fed3}
\end{figure}

\begin{figure}
\centering
\begin{subfigure}{1.0\textwidth}
  \centering
  \includegraphics[width=0.9\linewidth]{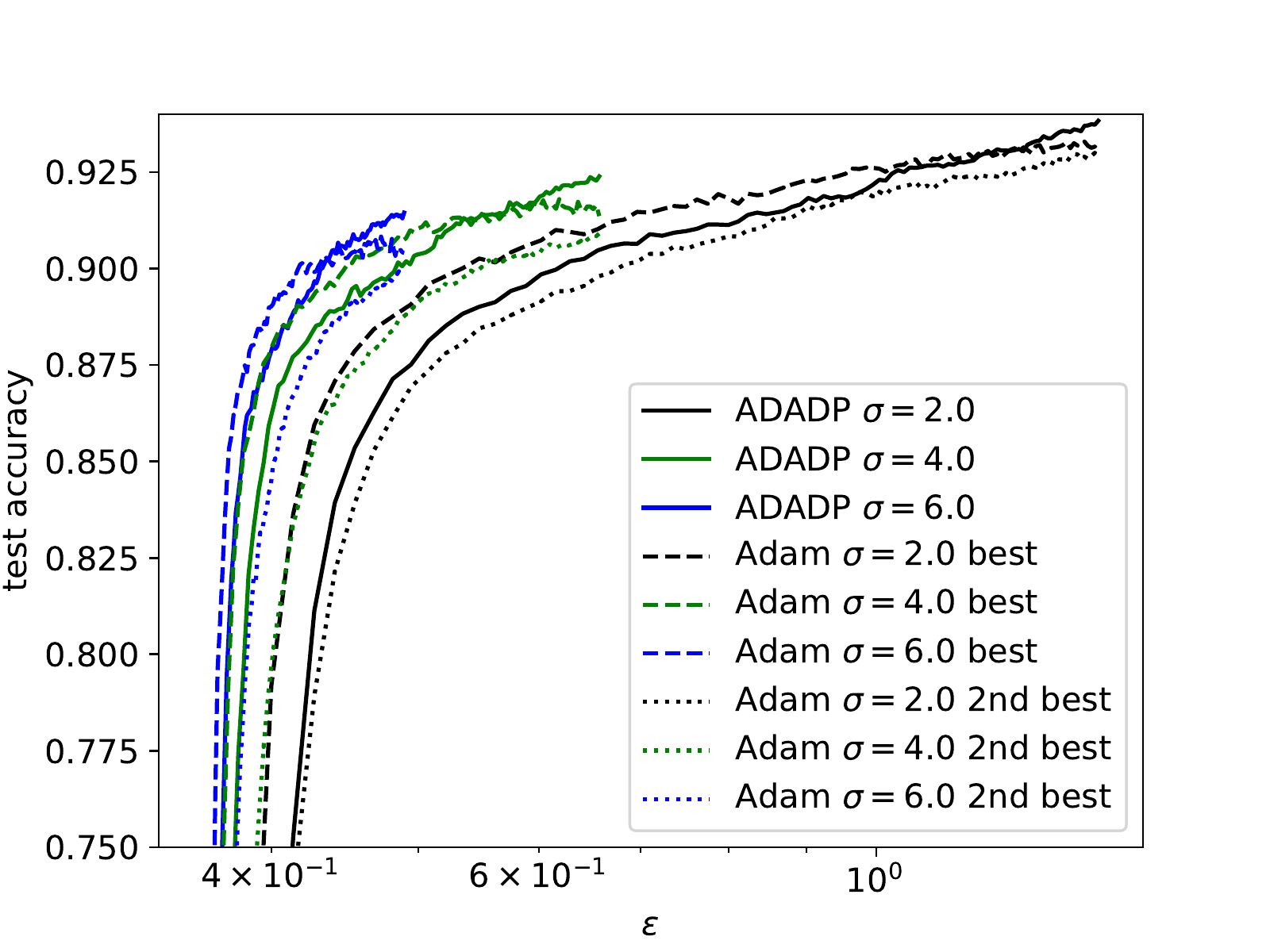}
  \caption{MNIST}
  \label{fig:mnist3}
\end{subfigure} \\
\begin{subfigure}{1.0\textwidth}
  \centering
  \includegraphics[width=0.9\linewidth]{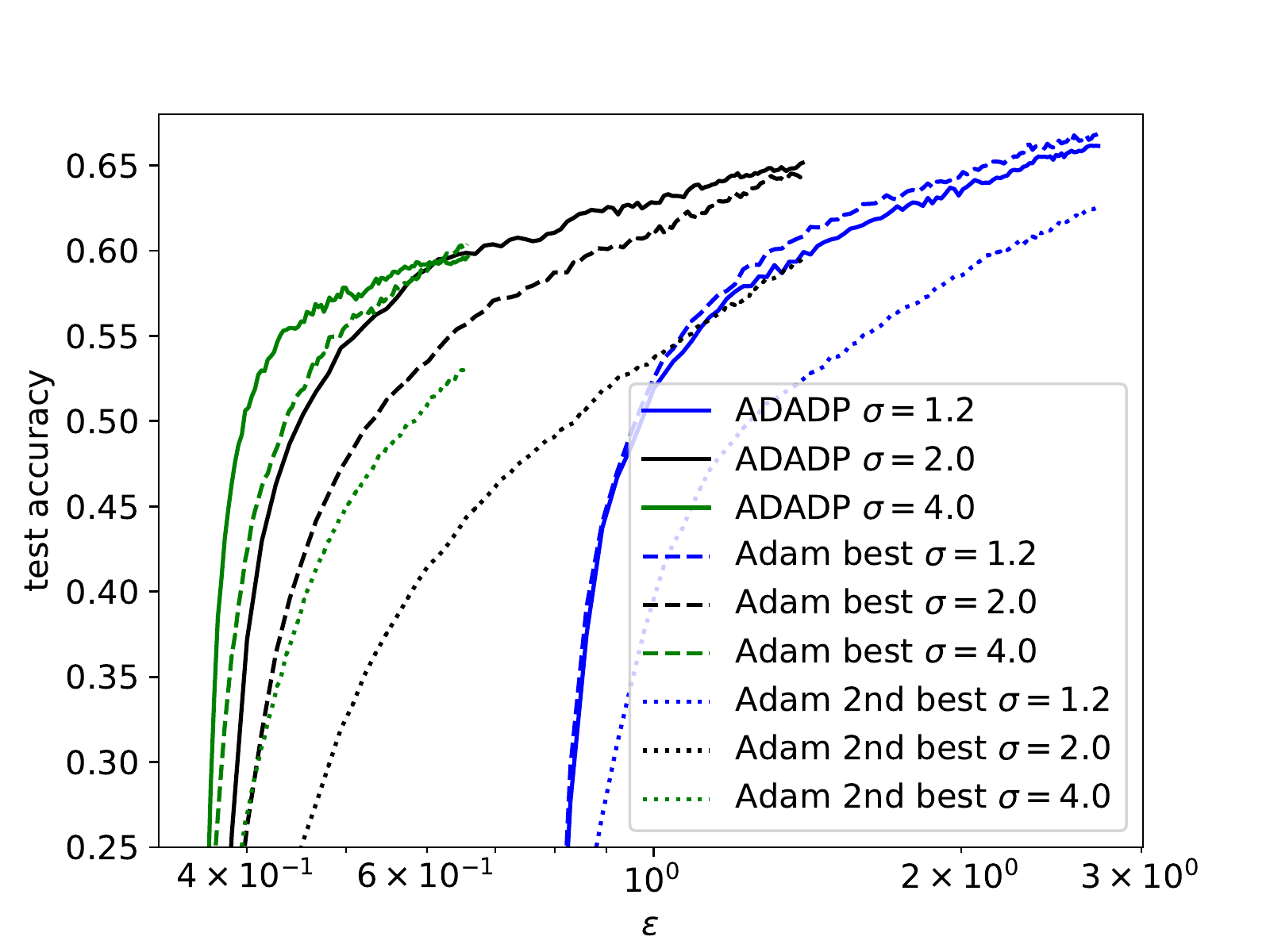}
  \caption{CIFAR-10}
  \label{fig:cifar2}
\end{subfigure}
\caption{DP learning using ADADP and Adam with optimal and nearly optimal initial learning rate $\eta_0$ sought from the grid $\{\ldots, 10^{-4.5}, 10^{-4}, 10^{-3.5}, \ldots\}$ for each $\sigma$.}
\label{fig:test}
\end{figure}
\newpage

 %%%%%%%%%%%%%%%%%%%%%%%%%%%%%%%%%%%%%%%%%%%%%%%%%%%%%%%%%%%%%%%%%%%%%%%%%%%%%%%%%%%%%%
%%%%%%%%%%%%%%%%%%%%%%%%%%%%%%%%%%%%%%%%%%%%%%%%%%%%%%%%%%%%%%%%%%%%%%%%%%%%%%%%%%%%%%
 \subsection{Comparison of ADADP against DP-Adam}
%%%%%%%%%%%%%%%%%%%%%%%%%%%%%%%%%%%%%%%%%%%%%%%%%%%%%%%%%%%%%%%%%%%%%%%%%%%%%%%%%%%%%%
%%%%%%%%%%%%%%%%%%%%%%%%%%%%%%%%%%%%%%%%%%%%%%%%%%%%%%%%%%%%%%%%%%%%%%%%%%%%%%%%%%%%%%

We use all the methods with minibatch size $\abs{B}=200$ and run each method for $100$ epochs.
The initial learning rate for ADADP is set to $10^{-1}$, but the results
are quite insensitive to this value as the algorithm will converge to
the desired learning rate already during the first epoch.
%Privacy parameter values $(\varepsilon,\delta)$ for different values of $\sigma$ are listed in Table~\ref{table:mnist}.

We first compare ADADP with optimally tuned Adam. This means that in each case
we search the best and the second best initial learning rate $\eta_0$ for Adam on a grid $\{\ldots, 10^{-4.5}, 10^{-4}, 10^{-3.5}, \ldots\}$.
%and then choose the second best $\eta_0$-value. 
We apply ADADP for 50 steps, then fix the learning rate (denoted $\eta_{50}$) and apply SGD with the decaying learning rate
$\eta_k = \tfrac{\eta_{50}}{1 + 0.1 \cdot(k-50)}$,
where $k$ denotes the number of epoch ($k>50$).

As Figure~\ref{fig:mnist3} illustrates, in case of MNIST and the feedforward network, ADADP is competitive with the learning rate optimised Adam
and gives better results than Adam with the second best learning rate found from the grid.
We see from Figure~\ref{fig:cifar2}, that in the case of CIFAR-10 and convolutional network, ADADP is again competitive with the learning rate optimised Adam
and gives clearly better results than Adam with the second best learning rate. % found from the grid.

%%
%%%%%%%%%%%%%%%%%%%%%%%%%%%%%%%%%%%%%%%%%%%%%%%%%%%%%%%%%%%%%%%%%%%%%%%%%%%%%%%%%%%%%%%%%%%%%%%%%%%%
%%%%%%%%%%%%%%%%%%%%%%%%%%%%%%%%%%%%%%%%%%%%%%%%%%%%%%%%%%%%%%%%%%%%%%%%%%%%%%%%%%%%%%%%%%%%%%%%%%%%
\subsection{Experiments for ADADP and SGD}
%%%%%%%%%%%%%%%%%%%%%%%%%%%%%%%%%%%%%%%%%%%%%%%%%%%%%%%%%%%%%%%%%%%%%%%%%%%%%%%%%%%%%%
%%%%%%%%%%%%%%%%%%%%%%%%%%%%%%%%%%%%%%%%%%%%%%%%%%%%%%%%%%%%%%%%%%%%%%%%%%%%%%%%%%%%%%

Next, we search an optimal learning rate for SGD on a grid $\{\ldots, 10^{-2.5}, 10^{-2.0}, $ \\ $10^{-1.5}, \ldots\}$
in the case $\sigma=2.0$. Using this learning rate for SGD, we compare the performance of SGD and ADADP when $\sigma=4.0$, $6.0$ and $8.0$.
As we see from Figures~\ref{fig:mnist3} and~\ref{fig:cifar2}, ADADP finds an appropriate learning rate and gives better results than SGD
for these values of $\sigma$.
This example is motivated by the fact that the learning rate found by ADADP is nearly constant after finding a suitable level.
Thus an optimally tuned SGD would necessarily be very competitive against ADADP.
One could expect to find a a suitable learning rate using the case $\sigma=2.0$.

%As we see from Figure~\ref{fig:cifar2}, ADADP is able to find an appropriate learning rate and gives better results when $\sigma=6.0,8.0$

%
\begin{figure}
\centering
\begin{subfigure}{1.0\textwidth}
  \centering
  \includegraphics[width=0.9\linewidth]{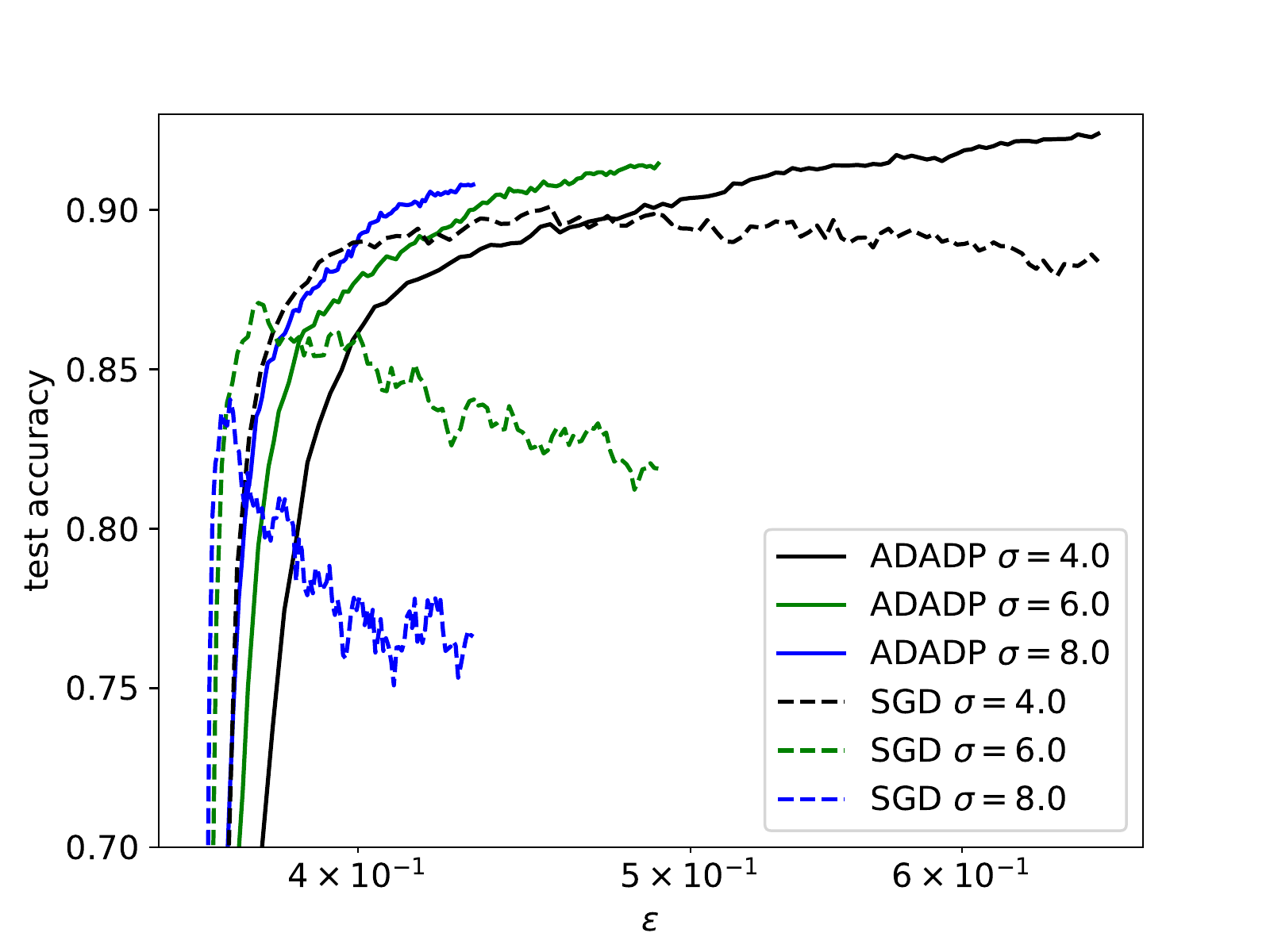}
  \caption{MNIST}
  \label{fig:mnist3}
\end{subfigure} \\
\begin{subfigure}{1.0\textwidth}
  \centering
  \includegraphics[width=0.9\linewidth]{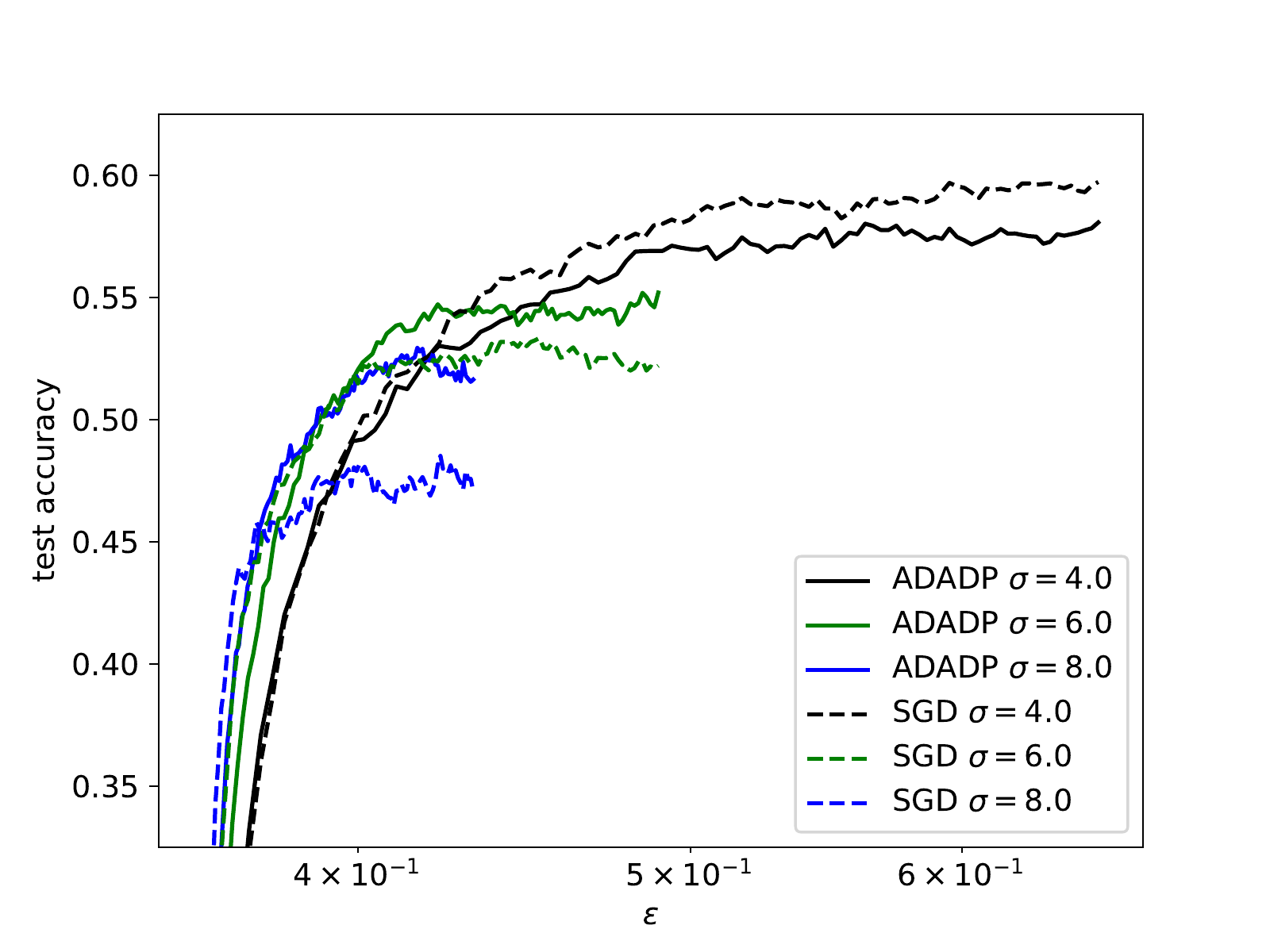}
  \caption{CIFAR-10}
  \label{fig:cifar2}
\end{subfigure}
\caption{ADADP and SGD. The fixed learning rate $\eta$ of SGD is tuned in the $\sigma=2.0$-case
 using the grid $\{\ldots, 10^{-2.5}, 10^{-2.0}, 10^{-1.5}, \ldots\}$.}
\label{fig:test}
\end{figure}

%%%%%%%%%%%%%%%%%%%%%%%%%%%%%%%%%%%%%%%%%%%%%%%%%%%%%%%%%%%%%%%%%%%%%%%%%%%%%%%%%%%%%%
%%%%%%%%%%%%%%%%%%%%%%%%%%%%%%%%%%%%%%%%%%%%%%%%%%%%%%%%%%%%%%%%%%%%%%%%%%%%%%%%%%%%%%
\section{Conclusions}
%%%%%%%%%%%%%%%%%%%%%%%%%%%%%%%%%%%%%%%%%%%%%%%%%%%%%%%%%%%%%%%%%%%%%%%%%%%%%%%%%%%%%%
%%%%%%%%%%%%%%%%%%%%%%%%%%%%%%%%%%%%%%%%%%%%%%%%%%%%%%%%%%%%%%%%%%%%%%%%%%%%%%%%%%%%%%

We have proposed the first learning rate adaptive DP-SGD method. We
believe this is the first rigorous DP-SGD approach, because
all previous works have glossed over the need to tune the SGD learning
rate.
%We have given rigorous moment bounds
%for the method, and using these bounds, we computed tight $(\varepsilon,\delta)$-DP bounds
%using the moments accountant.
By simple derivations, we have shown
how to determine the additional tolerance hyperparameter in the algorithm.
Based on this heuristic analysis, we developed a rule for selecting the
parameter and verified the efficiency of the resulting
algorithm in a number of diverse learning problems.
The results show that our approach is competitive in performance with
commonly used optimisation methods even without any tuning, which is
infeasible in the DP setting.
%This way, the user does not need to define the learning rate.
Overall, our work takes an important step toward truly DP and
automated learning for SGD-based learning algorithms.

Federated learning presents another setting where classical
hyperparameter adaptation with a validation set may be impractical and
also leads to suboptimal results. One obvious pain point is skewed
distribution of data on different clients, which may lead to different
clients requiring very different learning rates that would be very
difficult to tune without an adaptive algorithm. Our algorithm can
handle even highly pathological cases here with ease.

As a future work, it would be useful to develop a better understanding
of the tolerance hyperparameter.
Furthermore, it would be important to study the adaptation of
other key algorithmic parameters of DP-SGD, such as the gradient clipping
threshold and the minibatch size.
%, as especially the latter is
%closely linked with the learning rate.
\cite{Balles2017} provide an interesting
non-private implementation of minibatch adaptation, but unfortunately
their approach cannot easily be applied in the DP case.

\newpage

\bibliographystyle{plain}
\bibliography{adadp}

\end{document}